# NeuroSVM: A Graphical User Interface for Identification of Liver Patients

Kalyan Nagaraj[1*] and Amulyashree Sridhar[2]

[1*]PES Institute of Technology
kalyan1991n@gmail.com

[2]PES Institute of Technology
0908amulyashree@gmail.com

## Abstract

Diagnosis of liver infection at preliminary stage is important for better treatment. In today's scenario devices like sensors are used for detection of infections. Accurate classification techniques are required for automatic identification of disease samples. In this context, this study utilizes data mining approaches for classification of liver patients from healthy individuals. Four algorithms (Naïve Bayes, Bagging, Random forest and SVM) were implemented for classification using R platform. Further to improve the accuracy of classification a hybrid NeuroSVM model was developed using SVM and feed-forward artificial neural network (ANN). The hybrid model was tested for its performance using statistical parameters like root mean square error (RMSE) and mean absolute percentage error (MAPE). The model resulted in a prediction accuracy of 98.83%. The results suggested that development of hybrid model improved the accuracy of prediction. To serve the medicinal community for prediction of liver disease among patients, a graphical user interface (GUI) has been developed using R. The GUI is deployed as a package in local repository of R platform for users to perform prediction.

**Keywords** - Liver disease, SVM, Neural network, R, GUI

## 1. Introduction

Liver failures are at high rate of risk among Indians. The diagnosis of liver disorder at any early stage helps in designing effective prevention measures. In India, one in five persons is affected by liver disease. It is expected that by 2025 India may become the 'World Capital for Liver Diseases'. The widespread occurrence of liver infection in India is contributed due to deskbound lifestyle, increased alcohol consumption and smoking [1]. There are about 100 types of liver infections. Most prevalent liver infections are cirrhosis, non-alcoholic fatty liver disease and alcoholic liver disease. In the recent years sensors and mobile devices have been used extensively used for rapid detection of an infected liver [2]. The performance of these devices can be accelerated by spontaneous classification systems. Classification is a data mining technique comprising of a dual process flow. In the first step the classifier is trained using the training dataset, while the classifier is being tested for its prediction capacity in the second phase using different samples of the test set [3]. Feature selection is the preliminary step to be performed prior to application of classification algorithms for any dataset. There are three categories of feature selection techniques: filter based, wrapper based and embedded methods. Studies have depicted that the wrapper methods performs better compared to filter and embedded methods [4, 5].

Classification algorithms can be either supervised or unsupervised based on the learning mechanism. Supervised learning is implemented by set of labels defined prior in the training set. The function is mapped for new unseen data to predict the labels. Few examples are Discriminative learning, Artificial Neural Network, Bagging, Boosting, Naïve Bayes, Kernel-based classifiers, Nearest Neighbour algorithm, Decision Trees, Random Forest, and other ensemble of classifiers. Whereas unsupervised learning identifies the missing or hidden patterns in unlabelled data without any labels. They are commonly used for dimensionality reduction of feature space. The unsupervised ensembles include clustering approaches, self-organization maps (SOM), hidden Markov models (HMM) and adaptive resonance theory (ART) [7].

In this study, a hybrid NeuroSVM model is developed for classification of liver patients using Support Vector Machine (SVM) and Artificial Neural Network (ANN). Further, the hybrid model is implemented as a user friendly Graphical User Interface (GUI) in R. The GUI can be readily utilized by doctors and medical practitioners as a screening tool for liver disease.

2. **Literature Review**

Health care and medicine handles huge data on daily basis. This data comprises of information about the patients, diagnosis reports and medical images. It is important to utilize this information to decipher a decision support system. To achieve this it is important to discover and extract the knowledge domain from the raw data. It is accomplished by knowledge discovery and data mining (KDD) [8]. The implementation of data mining techniques is widespread in biological domain. Different studies have been conducted for classification of liver disorders, they are discussed briefly. In a study conducted by Rajeshwari et. al. different data mining algorithm like Naïve Bayes, FT Tree and KStar were implemented for analysis of liver disorder. It was found that Naïve Bayes algorithm resulted with an accuracy of 96.52% for the data [9]. In another study conducted by Ramana different classification algorithms were implemented for diagnosis of liver disorder. They include Naïve Bayes, C4.5, Neural Network, K-means and SVM. The performance of the algorithms was tested in different data sets. It was found that neural network classifier resulted in an accuracy of 71.59% for all the features in the liver dataset [10]. In another study conducted by Karthik et.al ANN was applied for classification followed by induction of rule set using Learn by Example (LEM) algorithm. It was followed by execution of fuzzy rules to identify different liver disease types which achieved an accuracy of 96% [11]. In another study the UCI liver dataset was used for selection of sub features based on random forest classifier with multi-layer perceptron induced [12]. Different artificial intelligence techniques were applied for the liver patient's dataset resulting in accurate predictions of liver malfunction [13, 14, 15, and 16].

Based on the review of literature, it was depicted that the past research studies have implemented different data mining techniques for classification of liver dataset. A hybrid model can be adapted to further increase the prediction accuracy of liver disease. It is followed by development of a graphical user interface would further aid the scientific community in early diagnosis of liver infection. It will provide a framework for end user application for generating promising treatment protocols.

3. **Methodology**

   3.1. **Data Collection**

   The Indian Liver Patient Dataset (ILPD) was selected from UCI Machine learning repository for this study. It is a sample of the entire Indian population collected from Andhra Pradesh region. The dataset comprised of 583 instances based on ten different biological parameters. The class value was reported based on these parameters as either yes (416 cases) or no (167 cases) to represent the liver infection.

   3.2. **Pre-processing and Feature selection**

   Pre-processing techniques was applied to normalize the missing values. The missing values along with their instances were replaced by null value. It was followed by feature selection to identify relevant attribute for classification. Feature selection was performed using both filter and wrapper methods. Initially the attributes having more than 70% correlation was removed from the dataset by correlation analysis. It was followed by feature selection using the library 'Boruta' in R. The algorithm was implemented on the basis of random forest for estimating the importance of different features in a dataset [17].

   3.3. **Randomization and splitting of dataset**

   The features selected in the preceding step were approved to develop classification models. Initially the dataset was randomized to obtain an arbitrary permutated sample. It was followed by splitting of the dataset into training (70% of the dataset) and test (30%) sets. Training set comprised of 389 instances and test set included the remaining 194 instances.

### 3.4. Classification algorithms

Different data mining algorithms like Naïve Bayes, Bagging, Random forest and Support vector machine (SVM) were implemented in R platform for classification. R is a popular statistical computing framework for performing data mining experiments. A 10-fold cross validation was performed for each of this algorithm. The algorithms are briefly discussed below:

**3.4.1.** *Naïve Bayes Classifier*: It is based on the Bayes theorem of conditional probability. The algorithm assumes that each attribute contributes to the total outcome independent of other attributes [18].

**3.4.2.** *Bagging*: They are applied for decision trees to reduce the rate of overfitting. Sampling and replacement is employed to generate a new training set from the given input training set. The bootstrap sample obtained is fitted with models using voting scheme for classification. The algorithm is considered as an averaging scheme for random decision trees. The algorithm was developed by Breiman [19].

**3.4.3.** *Random Forest*: They are classifiers that construct decision trees for training input. A random value is assigned as range to feature space for splitting the tree. Based on the training ensemble class value is predicted as the modal value of distinct tree. The algorithm was implemented by Breiman [20]. The algorithm is used for ranking the features by estimating out-of-bag error. It is followed by computing of important score for each feature.

**3.4.4.** *Support Vector Machine*: They are kernel based supervised classifier developed by Vapnik [21]. A hyperplane is constructed in high dimensional space defined by vector having a constant dot product. The feature space is mapped to the hyperplane of non-linear space using kernel function. The training optimized to the maximum margin hyperplane constitutes the support vectors. These vectors define an ideal linear hyperplane for classifying the instances.

### 3.5. Development of hybrid model

The hybrid model was developed with an aid to further increase the accuracy of liver disease prediction. Classification by SVM was figured out to be better compared to other algorithms. SVM was implemented in the hybrid model along with artificial neural network. A feed forward neural network was fed with the SVM classifier. The hybrid NeuroSVM predictor was implemented in R using the 'neuralnet' package [22]. Statistical significance of the hybrid model was evaluated using root mean square error (RMSE) and mean absolute percentage error (MAPE).

### 3.6. Deployment of hybrid model as GUI in R

The hybrid model was implemented as a user friendly GUI in R to facilitate the prediction of liver disease in a patient. The GUI was developed using 'gWidgets' 'RGtk2' and 'tcltk2' libraries in R [23, 24 and 25]. The GUI can be used as a screening tool for prediction of liver disease.

## 4. Results and Discussion

### 4.1. Dataset

The Indian Liver Patient Dataset comprised of 10 different attributes of 583 patients. The patients were described as either '1' or '2' on the basis of liver disease. The detailed description of the dataset is shown in Table-1.

| Sl. No | Attribute name | Attribute Type | Attribute Description |
|---|---|---|---|
| 1. | Age | Numeric | Age of the patient |
| 2. | Sex | Nominal | Gender of the patient |
| 3. | Total Bilirubin | Numeric | Quantity of total bilirubin in patient |
| 4. | Direct Bilirubin | Numeric | Quantity of direct bilirubin in patient |
| 5. | Alkphos Alkaline Phosphotase | Numeric | Amount of ALP enzyme in patient |
| 6. | Sgpt Alamine Aminotransferase | Numeric | Amount of SGPT in patient |
| 7. | Sgot Aspartate Aminotransferase | Numeric | Amount of SGOT in patient |
| 8. | Total Protiens | Numeric | Protein content in patient |
| 9. | Albumin | Numeric | Amount of albumin in patient |
| 10. | Albumin and Globulin Ratio | Numeric | Fraction of albumin and globulin in patient |
| 11. | Class | Numeric {1, 2} | Status of liver disease in patient |

**Table 1:** Description of Liver patient dataset

The methodology is represented as a flowchart in Fig 1.

Fig. 1 Flowchart for development of NeuroSVM model

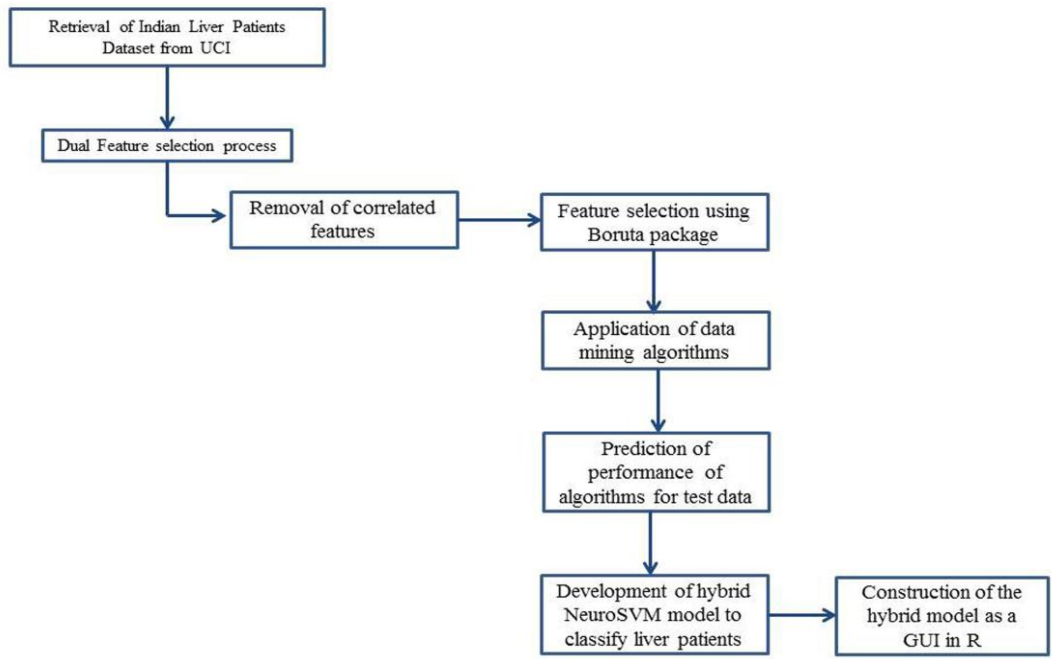

### 4.2. Feature Selection

Ten attributes available in ILPD dataset cannot be used to develop an accurate classification model. Only the relevant features must be selected. The features in dataset were subjected to correlation analysis for removal of redundant attributes. The correlated features were further subjected for feature selection using 'Boruta' package in R. It is a wrapper based feature selector algorithm that computes importance of each attribute by permutation. The attributes reported to be important after repeated iteration is chosen by the algorithm based on the ranks. The attributes selected by the algorithm is represented in Table-2. The plot derived from Boruta after feature selection is shown in Fig 2.

| Sl. No. | Attribute Name | Attribute Importance |
|---|---|---|
| 1. | Age | 0.9090 |
| 2. | Total Bilirubin | 1.000 |
| 3. | Direct Bilirubin | 1.000 |
| 4. | Alkphos Alkaline Phosphotase | 0.9504 |
| 5. | Sgpt Alamine Aminotransferase | 1.000 |
| 6. | Sgot Aspartate Aminotransferase | 1.000 |

**Table 2:** Attributes selected after dual selection

**Fig 2**: The selected attributes from Boruta (shown in green) after feature selection

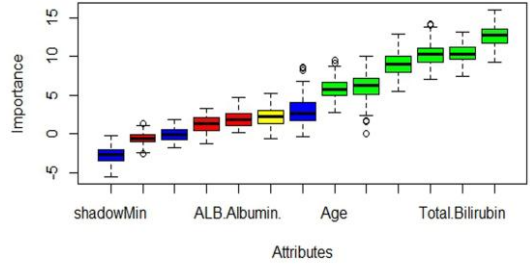

### 4.3. Application of classification algorithms

Four algorithms namely Naïve Bayes, Bagging, Random forest and SVM were implemented for classification of the Indian Liver Patient Dataset. The models were generated for the training set and evaluated on the test set. Based on the accuracy of prediction it was observed that SVM achieved an accuracy of 76.22%. The prediction outcome from SVM was used to develop a hybrid model.

### 4.4. Development of hybrid model

The hybrid model for assessment of liver diseases was build using neural network. The prediction results from SVM was fed into a feed forward artificial neural network comprising of 2 input layers, 5 hidden layers and one output layer. Neural networks were inspired from the neurons in the brain. The network is represented as a weighted graph with nodes and edges. The input function is represented by $x_j$ such that j=1, 2….n. The output function is computed using the formula:

$$y = \theta\left(\sum_{j=1}^{n} w_j x_j - u\right) \rightarrow \text{Equation (1)}$$

Here,
y= output function
$w_j$= weights applied for input j
$x_j$ = input
u= threshold function
θ= unit step function

The algorithm computes output class and network for input value based on the weights assigned. The algorithm was implemented using 'neuralnet' package in R. The hybrid model achieved a better performance compared to the individual algorithms. The performance of the prediction was estimated using the Receiver Operating Characteristics (ROC) curve. It is a plot of the false positive rate (FPR) versus the true positive rate (TPR). Root mean square error and mean absolute error was computed using the formula:

$$RMSE = \sqrt{\frac{1}{n}\sum_{i=1}^{n}(x_i - x)2} \rightarrow \text{Equation (2)}$$

$$MAPE = \frac{1}{n}\sum_{i=1}^{n}|x_i - x|/x_i \rightarrow \text{Equation (3)}$$

Here,
$x_i$= Actual class
x= Predicted class
n= Total number of instances

The value of RMSE was found to 1.164, predicting an overall error of 1.164%. It can be indicated as a reliable model for screening of liver patients. The accuracy of all the classification algorithms is shown in Table-3. The neural network plot for hybrid model is shown in Fig 3.

| Sl. No | Algorithm | Accuracy (%) |
|---|---|---|
| 01. | Naïve Bayes | 53.09 |
| 02. | Bagging | 66.73 |
| 03. | Random Forest | 67.67 |
| 04. | Support Vector Machine | 76.22 |
| 05. | NeuroSVM | 98.83 |

**Table 3:** Comparison of accuracy of classification algorithms

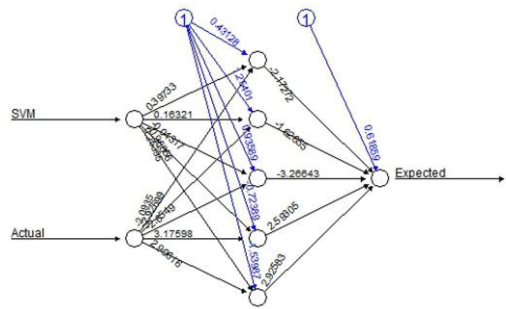

**Fig 3:** The NeuroSVM plot obtained from the hybrid model

### 4.5. Deployment of GUI in R

The hybrid NeuroSVM model is deployed as a GUI in R using packages 'gWidgets', 'RGtk2' and 'tcltk'. The biological attributes of patients is taken as input for prediction of liver disease. The attributes are subjected to prediction by SVM model. It is followed by implementation of feed forward hybrid NeuroSVM model for prediction of liver patients. The home page of the GUI is shown in Fig-4. Prediction of liver disease in the GUI using hybrid NeuroSVM is shown in Fig. 5 and Fig. 6. The GUI is further implemented as a package in R. The package 'NeuroSVM' is installed locally in the machine for implementation of liver disease prediction. The users can utilize the GUI by installing the package in R in their local machines. This application can be utilized by doctors and medicinal practisers to access the probability of liver infection in a patient for initial screening.

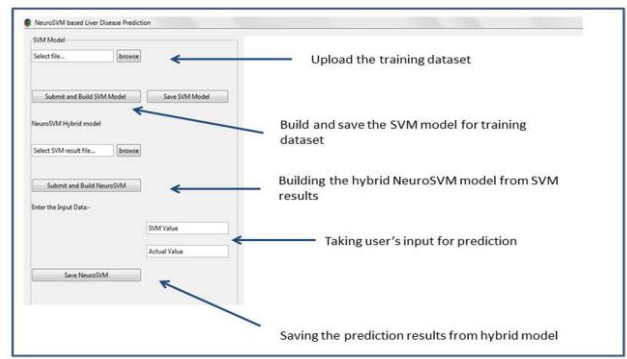

**Fig 4:** The home page of NeuroSVM GUI built in R

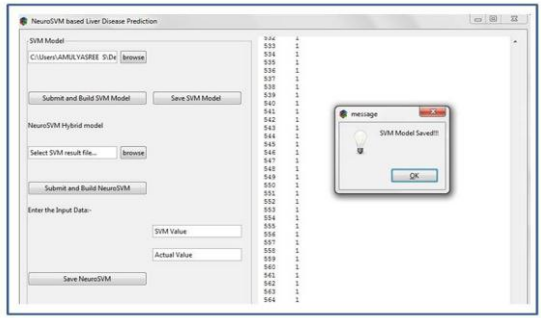

**Fig 5**: The prediction of SVM model in GUI for liver patients

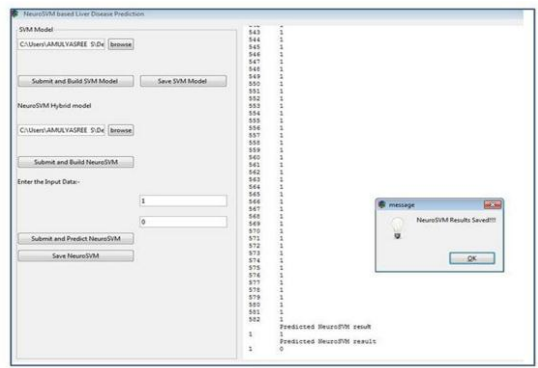

**Fig 6**: The prediction of NeuroSVM model in GUI for liver patients

## 5. Conclusion

Identification of liver infection at preliminary stage is important to combat the frequency and severity deaths of patients in India. The patients must be screened based on initial symptoms for development of personalized therapy. In this study, an attempt is made for prediction of liver disease in patients using data mining techniques. A hybrid NeuroSVM model was developed for classification of liver patients based on their biological parameters using artificial neural network. The hybrid model is deployed as a graphical user interface (GUI) in R. The GUI can be used as a screening tool by doctors for prediction of liver disease in patients in future.